\documentclass[conference]{IEEEtran}
\usepackage{color,amsfonts,lettrine,amsmath,amssymb,colortbl,booktabs,multirow,psfrag,bm,amsbsy,cite,multicol,lipsum,etoolbox}

\usepackage{relsize}

\newtoggle{2Collumn}
\toggletrue{2Collumn}
\makeatletter
\def\vect{\mathop{\operator@font vec}\nolimits}
\makeatother
\usepackage{algorithm,algcompatible}
\DeclareMathOperator*{\argmax}{\arg\!\max}% http://tex.stackexchange.com/q/83169/5764
\DeclareMathOperator*{\argmin}{\arg\!\min}% http://tex.stackexchange.com/q/83169/5764
\algnewcommand\INPUT{\item[\textbf{Input:}]}%
\algnewcommand\OUTPUT{\item[\textbf{Output:}]}%

\usepackage{enumerate}
\usepackage{subfigure}
\ifCLASSINFOpdf
   \usepackage[pdftex]{graphicx}
  \usepackage{epstopdf}
  \DeclareGraphicsExtensions{.pdf,.jpeg,.png}\else  % or other class option (dvipsone, dvipdf, if not using dvips). graphicx
	\DeclareMathOperator{\Tr}{Tr}
\fi

\DeclareMathOperator{\Tr}{Tr}
\begin{document}

\title{Deep Convolutional Neural Networks for Massive MIMO Fingerprint-Based  Positioning}

\author{
\IEEEauthorblockN{Joao Vieira, Erik Leitinger, Muris Sarajlic, Xuhong Li, Fredrik Tufvesson} \\
\IEEEauthorblockA{Dept. of Electrical and Information Technology, Lund University, Sweden \\
Joao.Vieira@eit.lth.se } }

\markboth{}%
{Shell \MakeLowercase{\textit{et al.}}: Bare Demo of IEEEtran.cls for Journals}

\maketitle

\begin{abstract}

This paper provides an initial investigation on the application of convolutional neural networks (CNNs) for fingerprint-based positioning using measured massive MIMO channels. When represented in appropriate domains,  massive MIMO channels have a sparse structure which can be efficiently learned by CNNs for positioning purposes. We evaluate the positioning accuracy of state-of-the-art CNNs with channel fingerprints generated from a channel model with a rich clustered structure: the COST 2100 channel model. We find that moderately deep CNNs can achieve fractional-wavelength positioning accuracies, provided that an enough representative  data set is available for training.

\end{abstract}


\section{Introduction }\label{sec:introduction}

Massive MIMO is a candidate technology to integrate next generation cellular systems, such as 5G systems, and deliver manifold  enhancements in the communications link  \cite{LUP4305564}. %Massive MIMO makes use of a  large number of base station (BS) antennas to serve a relatively low number of ubiquitous mobile terminals. 
In its  conceived form \cite{Marzetta}, massive MIMO  uses a  large number of base station (BS) antennas  together with \textit{measured} channel state information (CSI) to  multiplex user terminals spatially. Measured CSI is essential to yield spectrally efficient communications, but it can also be a key enabler to achieve highly-accurate terminal positioning,  where down to centimeter order accuracy may be required in some 5G applications, e.g., autonomous driving \cite{7885240}. Explained briefly, since positioning  is a spatial inference problem, it makes sense to use large antenna arrays that oversample the spatial dimension of a wireless channel (thus benefiting from, e.g., increased angular resolution, resilience to small-scale fading, and array gain effects) to aid the positioning task.%\footnote{Moreover, the ability of centralizing both communication and localization functionalities of a network into a single (and possibility co-located) large antenna array infrastructure is a convenient solution that  decreases the overall system deployment complexity and cost \cite{GuerraGD}. This contrasts with traditional deployments of dedicated systems for positioning  made out of several distributed anchor nodes.} %The research challenge is thus how to properly exploit measured channels for positioning.
% What makes positioning for massive MIMO a different problem than for point-to-point MIMO (non-planar wave assumption, non spatial stationarities, no effective channel fading).
%This makes massive MIMO  very suitable technology not only to integrate communication systems but also for positioning systems.

Several approaches that make use of measured massive MIMO channels for positioning  exist. For example, the approach proposed in \cite{7849233} detects a terminal's position (from a grid of candidate positions) using  line-of-sight (LOS) based triangulation from a terminal to several distributed massive MIMO BSs. In \cite{XuhongLi}, positioning is performed using  the phases of  multipath components estimated from massive MIMO channels.
 %Some limitations of the approach appear to be that: it always assumes the presence of LOS from the terminal to all BSs, and it  treats all remaining channel multipath components as non-informative for the positioning task. 
Another positioning approach was proposed in \cite{7390953}, where  received signal strength (RSS) based fingerprinting from one single-antenna terminal to  $N_{\rm BS}$ $M$-antenna BSs is employed. Here, the challenge was to learn  the inverse map
\begin{equation}
f_{\rm RSS}^{-1}: \{ | {\bf Y}_i|^2 \} \rightarrow \{ {\bf x}_i \}
\end{equation}
from a set of training observations, i.e. the training set $\{  {\bf Y}_i, {\bf x}_i \}_{i=1}^{N_{\rm Train}}$. Here the label ${\bf x}_i \in \mathbb{R}^{2 \texttt{x} 1 }$ is the 2-dimensional terminal coordinate of  training observation  $i$, and $ {\bf Y}_i \in \mathbb{C}^{M \texttt{x} N_{\rm BS}}$ is its associated channel fingerprint. Gaussian Process-based Regression was used to learn $f_{\rm RSS}^{-1}(\cdot)$. %The main positioning limitations of this approach appear to be: not accounting for the  phase information  availiable in the fingerprints set $\{  {\bf Y}_i \}_{i=1}^{N_{\rm Train}}$, and the  limited learning capability of Gaussian Kernels.

What the previous two mentioned positioning proposals and other proposals have in common (e.g., \cite{7684736} and \cite{7247265}),  is that the structure of their   solutions is typically composed by 2 distinct steps. In the first step, empirical feature extraction (from the measured channel snapshots)  is performed (e.g., RSS), and in the second step,   positioning of the terminal is done using the extracted features and a suitable algorithm-the algorithm typically being is the main contribution of the work. Although such 2 step solutions  simplify the entire positioning task, they are inherently sub-optimal since they are constrained to use only partial--and typically not statistically sufficient \cite{Kay}--channel statistics to solve the problem at hand. Thus, is it of interest to explore  positioning frameworks that jointly \textit{extract} and \textit{process} channel features from measurements--under some joint optimality criterion.

%and \textit{ii)} somewhat restrictive  assumptions on the wireless channel structure are made in order to either  choose relevant features to extract or to of analyse the proposal results, which calls the generality of the approach into question.

In this work, fingerprinting-based positioning is performed using a framework that jointly extracts and processes channel features for positioning. More specifically, we are interested to learn
\begin{equation}
\label{eq:Pos}
f^{-1}: \{ s\left(  {\bf Y}_i \right) \}  \rightarrow \{ {\bf x}_i \},
\end{equation}
i.e., the inverse of the underlying function $f(\cdot)$ that maps a set of single-antenna terminal coordinate vectors $\{ {\bf x}_i \}$ to their respective  \textit{measured} but transformed channel snapshots $ s\left(  {\bf Y}_i \right) \, \in \mathbb{C}^{d_1 \texttt{x} \dots \texttt{x} d_D} \; \forall i$, where $D$ is the dimensionality of the transformed snapshot. We note that the main point of the transformation $s(\cdot)$ is to  obtain a sparse representation for $ s\left( {\bf Y}_i \right)$. This is motivated in detail in Sec. \ref{sec:Motiv}. For now, we remark that the sparse transformations considered in this work are bijective, and thus yield no information loss.\footnote{We remark that this positioning approach is inherently  designed for the single-user case. This fits well within a massive MIMO context since mutually orthogonal pilot sequences, which are seen as sounding sequences in the context of this work, are typically used by different users during uplink training  \cite{Marzetta}. The extention of this approach to a multi-user case is thus straightforward.} % Here, $D$ is the number of channel dimentions, and $N_1$ and $N_2$ represent the sizes of the measured channel dimensions, e.g., $N_1$ can be the number of antennas of a 1-dimensional array, or the number of sampled delays of a wideband channel. %We remark that, once this map is learned, one simply needs to input a new measured channel to (\ref{eq:Pos}) to infeer the position at which it was measured at.

Our proposal to learn (\ref{eq:Pos}) is by means of deep convolutional neural networks (CNNs). Deep neural networks  provide state-of-the-art learning machines that yield the most learning capacity from all machine learning approaches \cite{Goodfellow-et-al-2016}, and lately have been very successful in image classification tasks. %Given that the measured snapshots ${\bf Y}_i$ have sparse structure when represented in appropriate domains, this problem resembles that of image classification where most relevant information for classification is condensed in sparse localitions of the image.
Just like most relevant information for an image classification task is sparsely distributed  at some locations of the image \cite{Goodfellow-et-al-2016}, measured channel snapshots ${\bf Y}_i$ have, when represented in appropriate domains, a sparse structure which--from a learning perspective--resemble that of images. This sparse channel structure can be learned by CNNs  and therefore used for positioning purposes. To the best of the authors' knowledge there is no prior work on this matter.%This is further motivated in Sec. \ref{sec:SysMol}.

The main contributions of this paper are summarized below.

\begin{itemize}
\item We investigate the feasibility of deep CNNs for fingerprint-based positioning with massive MIMO channels, and provide insights on how to design such networks  based on machine learning and wireless propagation theory.

\item As a proof-of-concept, we demonstrate the accuracy of our approach by performing fractional-wavelength positioning using channel realizations generated from a   widely accepted cellular channel model: the COST 2100 MIMO channel model \cite{6393523}.
\end{itemize}

%\begin{itemize}

%\item The positioning here is not based on high-resolution parameter estimation algorithm, tailored for signal processing of measured channels.  It can be used by realistic BS, to do online positining, and it accounts for the absolute phase accuracy. The approach is even more suited is some kind of classification needs to be done. A similar approach we do here is to oversample the input, using FFTs. However, this is a linear operation and can be inverted by the network if it find it suitable, if the first layer is a fully connected layer.

%\end{itemize}

\section{Channel Fingerprinting and Pre-Processing }
\label{sec:SysMol}

%In this section we describe the link setup considered in this work and its respective transformations of the inputs. We motivate why such input structure is suitable for CNNs, and provide remarks about on how to generalize the current link.

In this section, we explain the  fingerprinting scenario addressed in this work, and motivate why CNNs are  learning machines suitable to perform positioning  under such scenarios. To maximize insights, we focus most motivational remarks on the   current case-study, but also provide several generalization remarks at the end of the section.

\subsection{Channel Fingerprinting}
\label{sec:InpDimChMod}
In this work, we assume a BS equipped with a linear $M$-antenna  array made of omnidirectional $\lambda/2$-spaced elements, and that narrowband channels sampled at $N_F$ equidistant frequency points are used for positioning.
With that, the dimensionality  of  each channel fingerprint ${\bf Y}_i$ (and, as it will be seen later, of each transformed fingerprint s(${\bf Y}_i$)) is $D=3$ and $$d_1=M, \; d_2=N_F, \; {\rm and} \; d_3 = 2.$$
% 
% terminal position ${\bf x}_i$ takes the form
%\begin{equation}
%\label{eq:InputRep}
%{\bf Y}_i \in \mathbb{C}^{M \texttt{x} N_F}.
%\end{equation}

Given a terminal position, its associated fingerprint is generated through $f(\cdot)$, i.e. the inverse of the function we wish to learn. We implement $f(\cdot)$  using the COST 2100 channel model, the structure of which is illustrated by Fig. \ref{fig:Channel}, under the parametrization proposed in \cite{6410305}. This parametrization was performed for outdoor environments and is further detailed in Sec. \ref{sec:Res}. However, we note that our method is not restricted to work only in outdoor channels--we remark on the required channel properties in Sec. \ref{sec:Motiv}. It is important to note that, in this work,  $f(\cdot)$ is implemented as a bijective deterministic map, i.e., there is only one unique fingerprint per position.\footnote{Bijectiveness of $f(\cdot)$ applies in most practical propagation scenarios with high probability (the probability typically approaches one  as $M$ increases). This is an important aspect to consider as it addresses the conditions needed to be able to use CNNs (or more generally, to solve the inverse problem). On a different note, regarding the deterministic structure for $f(\cdot)$, this is done  by generating both training and test sets from the same given realization of the COST 2100 channel model stochastic parameters. This makes each fingerprint to be completely determined solely by the geometry of the propagation channel itself. Stochastic effects in the fingerprinting process, such as measurement and labeling noise, or even  time-variant channel fading are interesting impairments to be considered in the design of CNN  in future work. For now, we focus on the case of having  unique fingerprints per position, due to simplicity.} 

%As a remark, it is clear that with this setup, that there is only CNN needs to learn (\ref{eq:Pos}) per scenario.

%{\color[rgb]{1,0,0} Make clear that the map learned is for one particular scenario. }

%Each entry of (\ref{eq:InputRep}) is generated from the \textit{COST 2100} MIMO channel model - a  cellular channel model yielding many properties as in realistic channels.

%The We provide more details on the setup in .

\subsection{Motivation for CNNs and Sparse Input Structures}
\label{sec:Motiv}
Applying standard feed-forward neural networks to learn the structure of $\{{\bf Y}_i\}$  may be computationally intractable, specially when $M$ grows very large. However, the structure of neural networks can be enhanced, both from a computational complexity and a learning point-of-view, if designed with sparse interaction and parameter sharing  properties \cite{Goodfellow-et-al-2016}. This is a widely used architecture for CNNs, suitable to process inputs with grid-like structures (e.g., an image can be thought as a two-dimensional grid of pixels) with minimal amounts of pre-processing. 

\begin{figure*}
    \center
    \begin{tabular}{cc}
        \includegraphics[width=0.9\columnwidth]{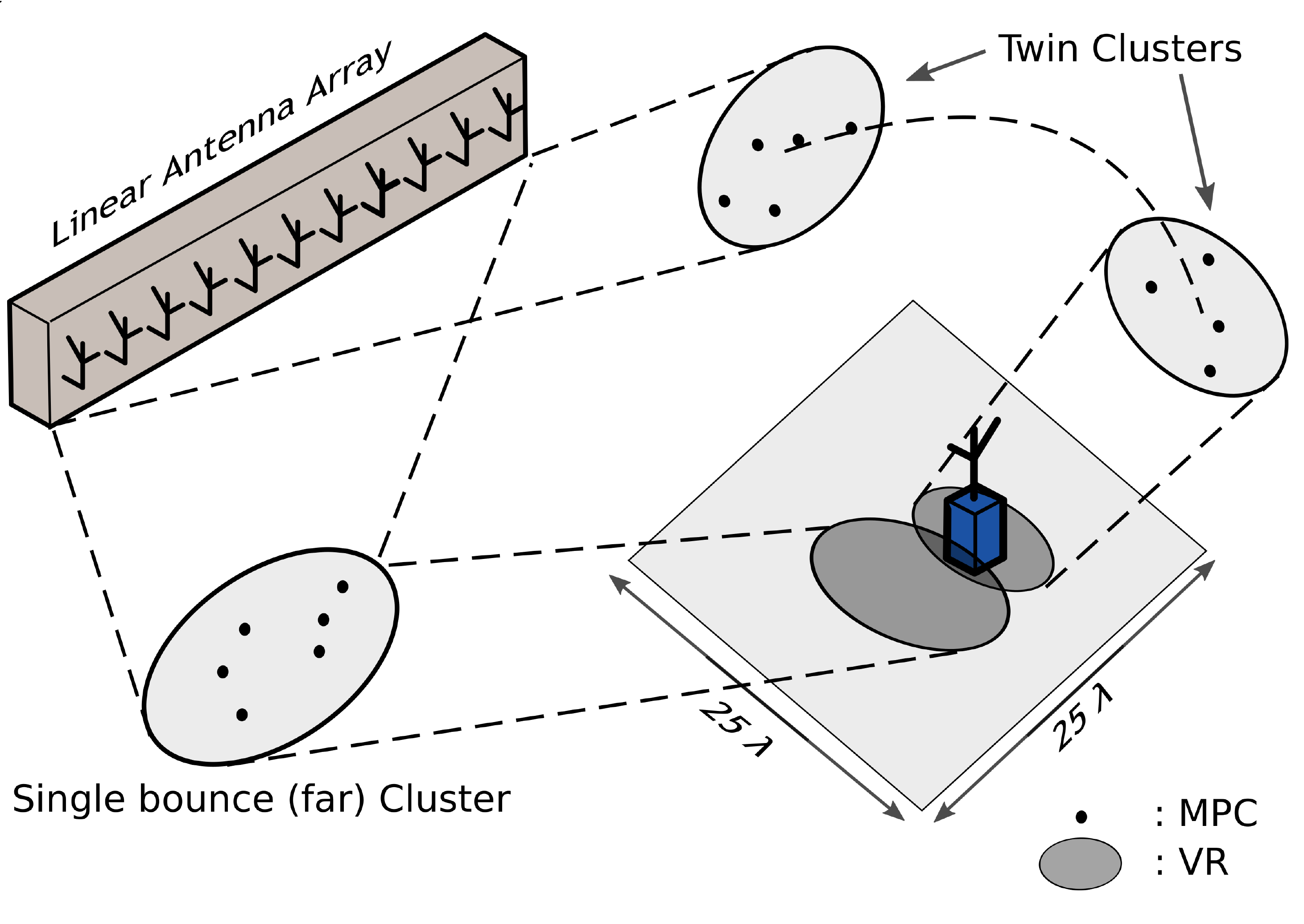}
        \includegraphics[width=1.15\columnwidth]{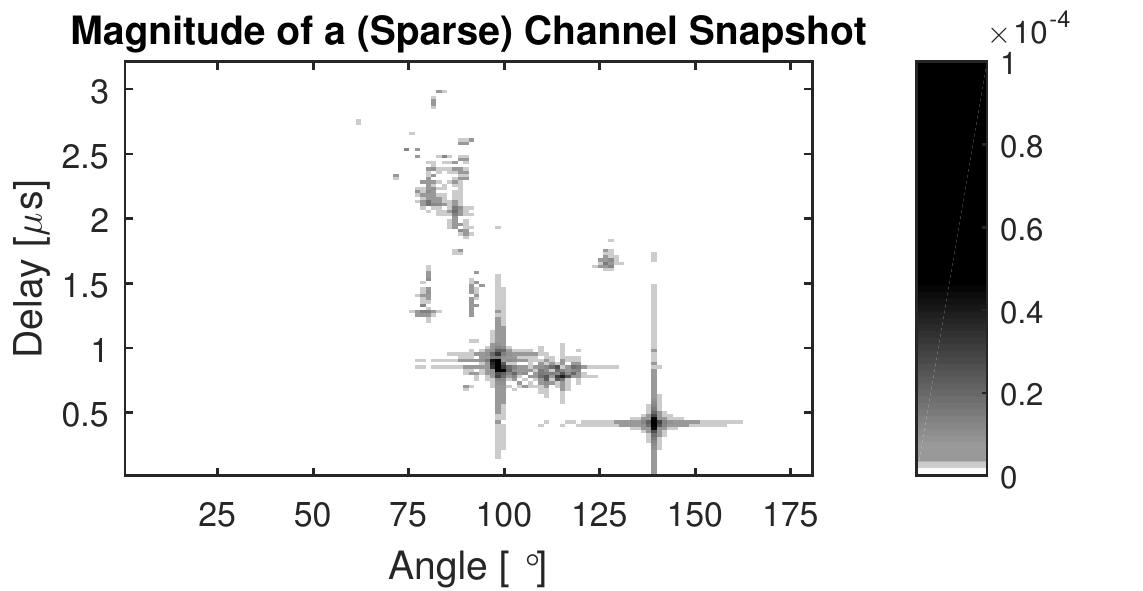}
    \end{tabular}
    \caption{Left--Link setup considered in this work: an $M$-linear BS array positioning one single-antenna terminal in a confined square area. Channel realizations are generated through the COST 2100 MIMO channel model. This geometry-based stochastic channel model is composed by different types of clusters of multipath components (MPCs) that illuminate certain visibility regions (VRs) of an area. Right--Example of the magnitude of a channel snapshot represented in a sparse domain. Such channel channel has a rich structure that can be learned by a CNN for positioning purposes.}
    \label{fig:Channel}
\end{figure*}

CNNs are efficient learning machines given that their inputs meet the following two 
structural assumptions:
\begin{enumerate}
\item most relevant information features are sparsely distributed  in the input space;
\item the shape of most relevant information features is invariant to their location in the input space, and are well  captured by a finite number of  kernels.
\end{enumerate}
From a wireless channel point-of-view, these assumptions apply well when channels snapshots (i.e., the CNN inputs)  are represented in domains that yield a sparse structure \cite{molisch2010wireless}. For example, in the current case study,   sparsity is achieved by representing ${\bf Y}_i$ in its, so-called, angular-delay domains, see Fig. \ref{fig:Channel}. Trivially,  $s(\cdot)$ can take the form of a two-dimensional discrete Fourier transform, i.e., 
\begin{equation}
\label{eq:Transf}
s({\bf Y}_i) = {\bf F} \, {\bf Y}_i  \, {\bf F}^H.
\end{equation}
%Due to the structure of (\ref{eq:InputRep}), this can be easilly achieved by a 2-dimensional Fourier transform of \ref{eq:InputRep}
%\begin{equation}
%\label{eq:InputTransf}
%\bar{{\bf Y}}_i = {\bf F}_M {\bf Y}_i {\bf F}_{N_T}^H,
%\end{equation}
%where ${\bf F}_n$ is the  discrete Fourier transform matrix of size $n$. 
If  specular  components of the channel, which are typically modeled by Dirac delta functions \cite{molisch2010wireless}, are seen as the information basis for positioning, then the two structural assumptions of the CNNs inputs listed above are met. The same  applies, if instead, clusters of multipath components are seen as the information features for positioning. 
%Under such channel representation, channel specular  components - which comprise most relevant information for positioning - are typically represented by Dirac delta functions  \cite{molisch2010wireless},  which therefore have a sparse structure that is invariant across the entire angular-delay spectrum.

\subsection{Generalizing the Current Case Study}
\label{sec:PreProSpI}
%The current case study focus on a simple setup, for simplicity, in order to be able to extract the main insights. However, o
The current case study can be  extended to more generic/higher-dimensionality fingerprints. For example, if $\{  {\bf Y}_i \}$ is comprised by snapshots measured from a multi-antenna terminal to a BS array, both with arbitrary array structures and non-omnidirectional antenna elements, the effective aperture distribution function (EADF) \cite{SmartAntennas} may be accounted in $s(\cdot)$ in order to obtain a sparse fingerprint representation. However, we emphasize that, in contrast to most propagation studies, one does not need to necessarily  de-embed the measurement system from the propagation channel in order to obtain valid fingerprints for positioning. Also, multi-antenna channels yielding phenomena such as violation of the plane wave assumption, or even the existence of cluster visibility regions \cite{6810277}, can be made sparse by means of proper transformations, e.g., generalized Fourier transforms.  In any case, the key is the ability to obtain a sparse representation for $ s\left({\bf Y}_i \right)$.

\section{Deep CNN Architecture}\label{sec:DCNNarq}
In this section, we describe the network architecture used for learning $f^{-1}(\cdot)$, and  discuss some design aspects.

For notation convenience, we drop the  dependence of the  training sample index $i$, and  write ${\bf Y} \triangleq {\bf Y}_i$ and ${\bf x} \triangleq {\bf x}_i$ until explicitly stated otherwise.

\subsection{Convolutional-Activation-Pooling Layers}
After the input layer, which takes the transformed snapshots, a typical structure of CNNs employs a cascade of $L$ convolutional-activation-pooling (CAP) layers. Each CAP layer is composed by: \textit{i)} a convolutional operation of its input with  $K$ convolutional Kernels, \textit{ii)} a  non-linear transformation, i.e., activation function, and \textit{iii)} a pooling layer, respectively. A detailed description  of the CAP layer structure used in  this work  follows below.

%Deep networks are characterized by having $L$ large.% A short description  of the network used is provided next.

%Let ${\bf y}_{r,c}$ be a submatrix of (\ref{eq:InputTransf}) starting at the $r$th row and $c$th collumn, and the $j$th convolutional Kernel, with $1 \leq j \leq K_n$,  ${\bf w}_j \in \mathbb{C}^{s_1 \texttt{x} s_2 \texttt{x} s_3}$ with $s_1<<M$ and $s_2<<N_F$. 
Let the tensor ${\bf H}^{\ell -1} \in \mathbb{R}^{M \texttt{x} N_F \texttt{x} S_3}$ be the input of the $\ell$th CAP layer,  with $1 \leq \ell \leq L$. Also, let  the $j$th convolutional Kernel   of the $\ell$th layer be denoted by ${\bf w}^\ell_j \in \mathbb{R}^{S_1 \texttt{x} S_2 \texttt{x} S_3}$, with $1 \leq j \leq K$, and $S_1$   and $S_2$ denoting the sizes of the Kernels (which are CNN hyper-parameters). With ${\bf h}^{\ell -1}_{r,c} \in \mathbb{R}^{S_1 \texttt{x} S_2 \texttt{x} S_3}$ being a sub-tensor of a zero-padded version\footnote{The zero-padded version of ${\bf H}^{\ell -1}$ is obtained by padding the borders of the volume of ${\bf H}^{\ell -1}$ with zeros, such that, when convolved with any Kernel ${\bf w}^\ell_j$, the input and output volumes are the same \cite{Goodfellow-et-al-2016}.} of ${\bf H}^{\ell -1}$, an output entry of the ${\ell}$th convolutional layer can be written as
%Let ${\bf y}_{r,c}$ be a submatrix of (\ref{eq:InputTransf}) starting at the $r$th row and $c$th collumn, and the $j$th convolutional Kernel, with $1 \leq j \leq K_n$,  ${\bf w}_j \in \mathbb{C}^{s_1 \texttt{x} s_2 \texttt{x} s_3}$ with $s_1<<M$ and $s_2<<N_F$.
\begin{equation}
\label{eq:Conv}
c_{r,c,j}^\ell =   b_j^\ell +   {\bf 1}^T \left({\bf w}_j^\ell \circ {\bf h}^{\ell -1}_{r,c} \right) {\bf 1}.
\end{equation}
Here $\circ$ denotes the Hadamard product, $b_j^\ell$ is a bias term, $ {\bf 1} $ denotes the all-ones column vector, and $r$ and $c$   are indices in the convolution output volume which are implicitly defined.

In the first CAP layer,  the  input  ${\bf H}^{0}$ is a tensor made out of the complex-valued entries of (\ref{eq:Transf}), and thus we have  $S_3 = 2$ (real dimensions). This is because, although channel snapshots are inherently complex-valued, we pursue an implementation of a real-valued CNN with real-value inputs (thus we have $S_3 = 2$ in the first CAP layer). We motivate why we do so in Sec. \ref{sub:NetDes}. In the remaining CAP layers, we have $S_3 = K$.

Each convolutional output entry (\ref{eq:Conv}) is fed to an activation function. We use the current default choice for  activation functions in CNNs, namely, the  rectified linear unit (RELU) \cite{Goodfellow-et-al-2016}, where the output can be written as
\begin{equation}
\label{eq:RELU}
g_{r,c,j}^\ell={\rm max}\left( c_{r,c,j}^\ell, \;  0 \right).
\end{equation}
Finally, after each activation function follows a pooling operation which down samples the outputs of the activation functions.  A standard option, also used here, is to forward propagate the maximum value within  group of $N_1 \times N_2$ activation functions outputs. The pooling result  can be written as
\begin{equation}
h_{r,c,j}^\ell =  \underset{m=1 }{\stackrel{N_1}{\rm{max}}} \;\;  \underset{n=1 }{\stackrel{N_2}{\rm{max}}}\left(   g_{(r-1) N_1+m , \; (c-1)N_2+n, \; j}^\ell \right).
\end{equation}

\subsection{Fully-Connected Layer}

A fully-connected layer, following the  $L$ CAP layers,  finalizes the CNN. With that, the position estimate of the network, ${\bf t} \in \mathbb{R}^{2 \texttt{x} 1}$, is given by
\begin{equation}
\label{eq:FLC}
{\bf t} = {\bf W} \vect \left\{  {\bf H}^{L}  \right\}  + {\bf b}^L,
\end{equation}
where $\vect\left\{ \cdot \right\}$ vectorizes its argument, ${\bf b}^L = [b_1^L \; b_2^L]^T$ is a vector of biases, and ${\bf W}$ is a weight matrix whose structure is implicitly defined.
\subsection{Network Optimization  }

The CNN network learns the weights ${\bf W}$ and $\left\{ {\bf w}_j^\ell \right\}$, and biases $\left\{ b_j^\ell \right\}$, in order to make ${\bf t}$ the best approximation of $\bf x$. Since we address positioning as a regression problem, we use the squared residuals averaged over the training set as the optimizing metric. Re-introducing the dependence on the sample index $i$, and defining the column vector $ \boldsymbol \theta$ by stacking all network parameters as ${\boldsymbol \theta}=\left[ \vect\left\{[{\bf w}_1^1 \dots  {\bf w}_L^K ]\right\}^T  \vect\left\{  {\bf W}  \right\}^T [  b_1^1 \dots b_2^L ] \right]^T$, the optimum parameters are given by
\begin{equation}
\label{eq:MSE}
\hat{\boldsymbol \theta}   =  \underset{{\boldsymbol \theta} }{\argmin} \;\; J({\boldsymbol \theta}),
\end{equation}
with
\begin{equation}
\label{eq:Cost}
J({\boldsymbol \theta})  =  \frac{\beta}{2} {\boldsymbol \theta}^T{\boldsymbol \theta} + \frac{1}{N_{\rm train}} \sum\limits_{i=1}^{N_{\rm train}} ({\bf x}_i - {\bf t}_i({\boldsymbol \theta} ))^2.
\end{equation}
A Tikhonov penalty term is added to harvest the benefits of regularization in CNNs--$\beta$ is its associate hyper-parameter. On a practical note, we minimize (\ref{eq:Cost}) using stochastic gradient-descent and back propagation \cite{Goodfellow-et-al-2016}.

\subsection{Network Design Considerations}
\label{sub:NetDes}

We now motivate our choice for the implementation of real-valued CNNs. From our experience, the main challenge of generalizing CNNs to the complex-valued case appears to be in finding a \textit{suitable} generalization of (\ref{eq:RELU}), the RELU, in order to "activate" complex inputs. For example, the complex-valued CNN generalizations presented in \cite{Guberman16} apply (\ref{eq:RELU}) to both real and imaginary parts separately, which from our experience appears not to perform well from a network optimization point-of-view. Explained briefly, such RELUs are non-continuous functions in $\mathbb{C}$, as opposed to  only being non-differentiable in $\mathbb{R}$ as in real-valued network. This makes the network's optimization unstable with the current optimization method (i.e. gradient descent). Therefore, as real-valued CNNs have been very successful in image classification tasks, we conservatively choose to pursue this design option, and let the extension to the complex-valued case to be a matter of future work.

A practical remark regarding the choice of the number of CAP Kernels, $K$, follows. In practice, information features, addressed in Sec. \ref{sec:Motiv}, are subject to a number of variability factors. For example, the shape of a "\textit{measured} Dirac delta function" can vary with its location due to  discretization. Also, it is clear that, there is a higher variability in the informational features if they are seen as clusters of MPCs, rather than single MPCs themselves. In both cases, a practical approach to deal with such variability factors is to set a high number for $K$, and let the CNNs learn a set of kernels that span most possible variations.

To finalize, if the information features are seen as the clusters of MPCs, the sizes of the Kernels, $S_1$ and $S_2$, should cover their  range  in the angular and delay channel representations. This is because  MPCs within a cluster may be statistically dependent, but different clusters  are typically not.

\subsection{Complexity Aspects}

The most computationally challenging aspect of the entire approach is the  optimization stage of the network. This is due to the large size of the training sets, network dimensionality, non-convexity of $J({\boldsymbol \theta})$, etc. However, once  the optimization stage is finalized, \textit{real-time} positioning can easily be  achieved  due to the feed-forward structure of the network. This can be easily observed (in the current case-study) by looking at the overall complexity order of a CNN point-estimate:  $\mathcal{O}(K^2 M L N_{F} S_1 S_2)$. The fact that the complexity does not depend on the training set size $N_{\rm train}$ is one main advantage of using CNNs for positioning.

\section{Positioning Results}\label{sec:Res}

Next, we address the positioning capabilities of CNNs by means of numerical results. We omit showing optimization aspects of the network (e.g., convergence across epochs) as the main point of the paper is to analyse the positioning capabilities of an optimized CNN.

\subsection{Simulation Setup}
\label{SimulationSetup}

The setup used in our experiments is illustrated by Fig. \ref{fig:Channel}: the terminal is constrained to be in a square area $\mathcal{A}$ of $25 \times 25$  wavelengths.
Channel fingerprints are obtained in this area through the COST 2100 channel model under the $300$\,MHz parameterization (e.g., for path-loss and cluster-based parameters) established in \cite{6410305}. The remaining parameters are shown in 
Table \ref{table:SystemParam}, and the other CNNs hyper-parameters, i.e. $L$ and $K$, are varied during the simulations. 

\small
\begin{table}[ht]
    \caption{Channel and CNNs parameters}
    \noindent\begin{tabular*}{\columnwidth}{@{\extracolsep{\stretch{1}}}*{7}{l}@{}}
        \toprule
        \textbf{Parameter} & \textbf{Variable} & \textbf{Value} \\
        \midrule
        Carrier frequency 	&  $f_c$ 	& 300\,MHz \\
        Bandwidth 			& $W$ 		& 20\,MHz \\
        \# Frequency points 	&$N_F$ & 128 \\
        \# BS antennas & $M$ & 128 \\
        First BS antenna coordinate & ${\bf B}_1 $ & $[-200 \lambda \; \; -200 \lambda]^T$ \\
        Last BS antenna coordinate & ${\bf B}_M $ & $[-200 \lambda \; \; -200\lambda+\frac{(M-1)\lambda}{2}]^T$ \\
        %							\# antennas per UE & $A_{UE}$ & 1 \\
        \midrule
        Tikhonov hyper-parameter & $\beta$ & $10^{-3}$ \\
        Kernel angular length $[\;^\circ]$			& $S_1$  & $9.8$ \\
        Kernel delay length	$[\mu s]$		& $S_2$  & $0.175$ \\
        Pooling windows length 	& $N_1 \, {\rm and} \, N_2 $  & $2$ \\
        \bottomrule
    \end{tabular*}
    \label{table:SystemParam}
\end{table}

\normalsize
The closest and furthest  coordinate points of $\mathcal{A}$ with respect to the first BS antenna are: $${\bf u}_c =[-12.5\lambda \; -12.5\lambda]^T \; \textrm{and} \; {\bf u}_f =[12.5\lambda \; 12.5\lambda]^T,$$ respectively (i.e., the user is at least $||{\bf u}_c-{\bf B}_1||/\lambda$ wavelengths away from the first BS antenna). The coordinates of these two spatial points implicitly define the relative orientation of the linear array with the area $\mathcal{A}$. Similarly, the upcoming performance analysis is done by means of the normalized root mean-squared error (NRMSE) where the mean  is calculated as the average over the test sets samples. Thus, we have $${\rm NRMSE} =\frac{1}{\lambda } \sqrt{\frac{1}{ N_{\rm test}}   \sum\limits_{i=1}^{N_{\rm test}} ({\bf x}_i - {\bf t}_i({\boldsymbol \theta} ))^2}.$$ This error metric has an understandable physical intuition as it shows how the error distance relates to the wavelength.

%        \# User distribution for test error & $p(x,y)$ & $[\,\mathcal{U}(200,225) \, \, \mathcal{U}(200,225)\,]^T$ \\
%        \# User position grid for trainig & $u$ & $[100 \, 125]^T$ \\

The CNN training and testing is described as follow:
\begin{enumerate}
\item First, the training set is obtained by fingerprinting  a 2-dimensional uniformly-spaced (thus, deterministic) grid of positions spanning the totality of $\mathcal{A}$. The impact of the sampling density is discussed in Sec. \ref{sec:TraidngGrids}.
\item For the test set, each fingerprint's position is obtained by sampling a random variable with a uniform distribution with support $\mathcal{A}$.
\end{enumerate}

Note that, if the CNN cannot use the available fingerprints  for training, then the position estimator is $\mathbb{E}\{\bf x\}=0$, see (\ref{eq:Cost}). Its NRMSE, for the current case study, is given by
\begin{equation}
\label{eq:REF}
{\rm NRMSE^{ref}}(\mathcal{A}) = \frac{1}{\lambda } \sqrt{ \frac{1}{ \int_{\mathcal{A}} \partial {\bf d}   } \int_{\mathcal{A}} \left( {\bf d}  -  \mathbb{E}\{{\bf x}\} \right)^2 \, \partial {\bf d} }  \approx 10.2.
\end{equation}
Obviously, this reference value increases when $\mathcal{A}$ is larger. Since an optimized CNN with a non-zero number of fingerprints should be able to do better or equal than (\ref{eq:REF}),  we  use (\ref{eq:REF}) as a reference level in the analysis.

For benchmarking purposes, we also contrast our CNN results against a standard non-parametric fingerprinting approach \cite{7029113}. Seeing a training fingerprint as a function of its position, i.e. ${\bf Y}_{i}({\bf x}_{i})$, this approach computes the position from a new fingerprint ${\bf Y}_{\rm new}$ through a grid-search over normalized correlations as
\begin{equation}
\label{eq:ML}
\hat{{\bf x}}_i   =  \underset{ {\bf x}_i \in \{{\bf x}_i \}_{i=1}^{N_{\rm train}}    }{\argmax} \;\; \frac{ |\Tr\{  {\bf Y}_{i}({\bf x}_{i})^H {\bf Y}_{\rm new} \}|}{ \sqrt{  |\Tr\{  {\bf Y}_{i}({\bf x}_{i})^H  {\bf Y}_{i}({\bf x}_{i}) \} \Tr\{ {\bf Y}_{\rm new}^H {\bf Y}_{\rm new} \}  |   }   }.
\end{equation}
The following remarks can be made about this non-parametric approach:
\begin{enumerate}
	\item Compared to the use of CNNs, a main drawback is its computational complexity order, $\mathcal{O}(M  N_{F}^2 N_{\rm train})$, as it depends on the size of the training set;
	\item  it has no inherent interpolation abilities, and thus its error can be lower bounded given the spatial density of the training set;
\end{enumerate}

\subsection{Proof-of-Concept and Accuracy for Different CNN Parametrizations}

%\begin{figure}
%        \includegraphics[width=1.05\columnwidth]{./figures/Lund.pdf}
%    \caption{Lund Movement. Training set of $\lambda/4$}
%    \label{fig:LundMov}
%\end{figure}

\begin{figure*}[t]
    \center
    \begin{tabular}{cc}
            \includegraphics[width=1\columnwidth]{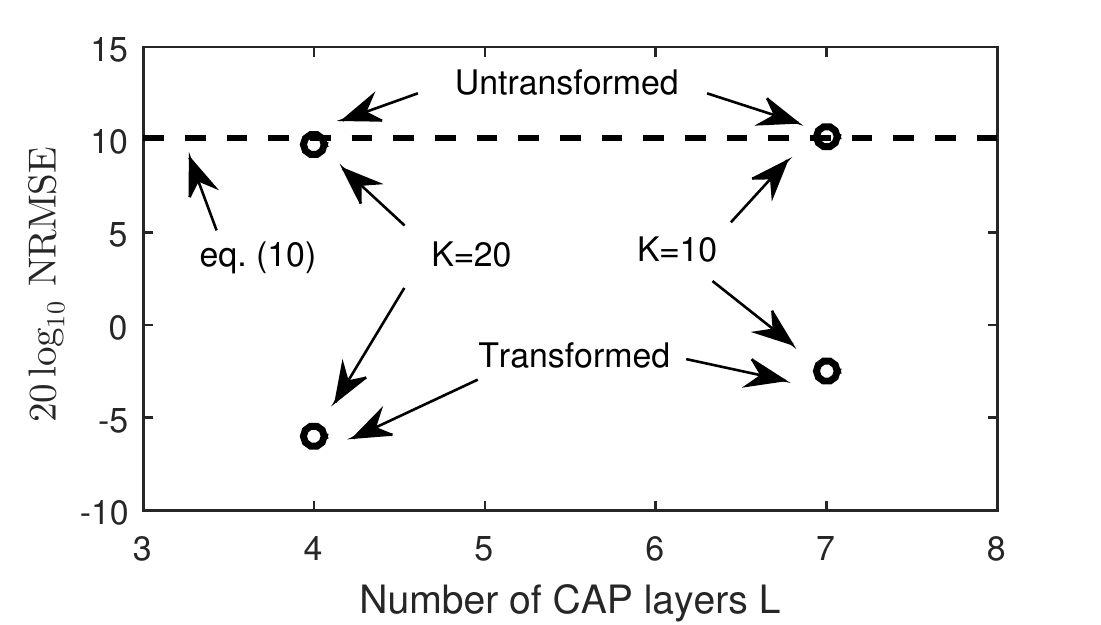}
            \includegraphics[width=1\columnwidth]{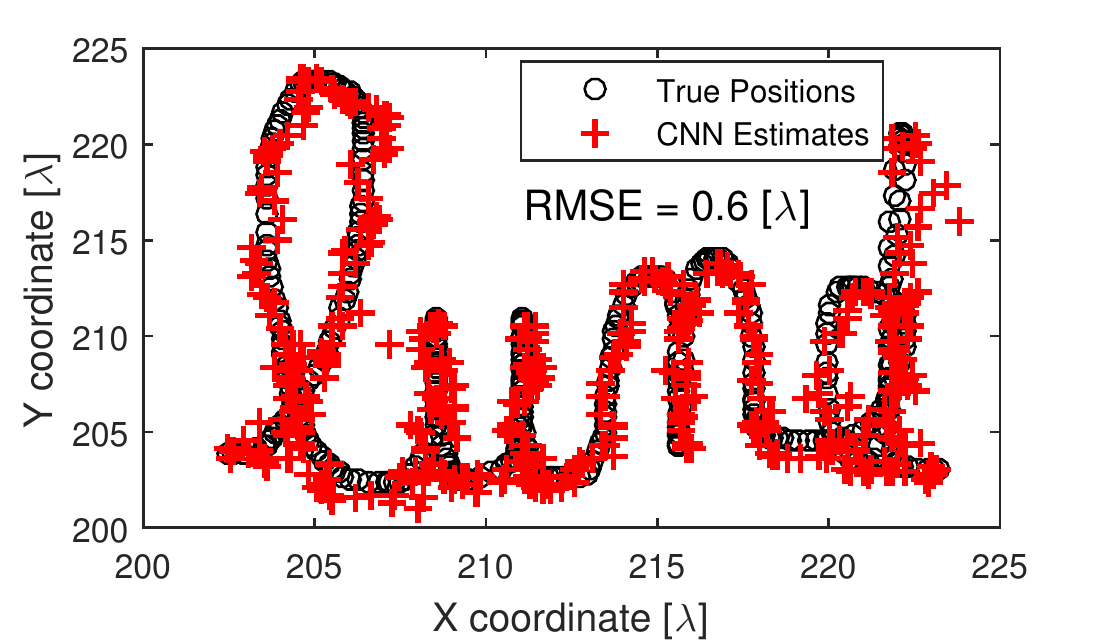}
    \end{tabular}
    \caption{Left--NRMSE obtained by CNNs under different parameterizations. The upper horizontal line corresponds to the reference level  (\ref{eq:REF}). Here we only report the test error, since a similar error value was obtained during training (i.e., no overfitting exists). Right--An illustrative example of the point-estimates from a pre-defined set of positions by the optimized CNN (from the left figure) with $K=20$ and $L=4$.}
    \label{fig:MSEpos}
\end{figure*}

Here, we report the positioning results when the spacing between neighbor training  fingerprints is $\lambda/4$. We consider this extreme case for now, in order to mitigate the impact of spatial undersampling from the results-here the focus is solely on the positioning capabilities of the network. The impact of the training fingerprints spacing is addressed later in Sec. \ref{sec:TraidngGrids}.

Fig. \ref{fig:MSEpos} (left) illustrates the positioning accuracy for different cases of CNN parameterizations. First, and as a sanity check, we see that for the same  parameterizations, a network fed with untransformed inputs (i.e., $s\left(  {\bf Y}_i \right) = {\bf Y}_i$) cannot effectively learn the channel structure for positioning purposes--the order of magnitude of the positioning error is similar to (\ref{eq:REF}). However, with transformed inputs, fractional-wavelength positioning can be achieved in both network settings, with the lowest achieved test NRMSE being of about $-6{\rm dB}\approx 1/2$ of a wavelength.  This showcases the capabilities of CNNs to learn the structure of the channel for positioning purposes. We remind that such positioning accuracies  are attained with only $20$ MHz of signaling bandwidth, see Table \ref{table:SystemParam}, which suggests that CNNs can efficiently trade-off signal bandwidth by BS antennas and still achieve very good practical performance. Decreasing the error further than fractional-wavelength ranges  becomes increasingly harder due to the increased similarities of nearby fingerprints--such range approaches  the coherence distance of the channel.  Also, as an illustrative example of the positioning accuracy, Fig. \ref{fig:MSEpos} (right)  shows the point-estimates of the CNN under the parameterization that attained the lowest test NRMSE. 
%{\color[rgb]{1,0,0} Maybe I can make a plot with the similarities of COST fingerprints as a function of distance.}

\subsection{Accuracy for Different Training Grids}
\label{sec:TraidngGrids}

\begin{figure}
        \includegraphics[width=1\columnwidth]{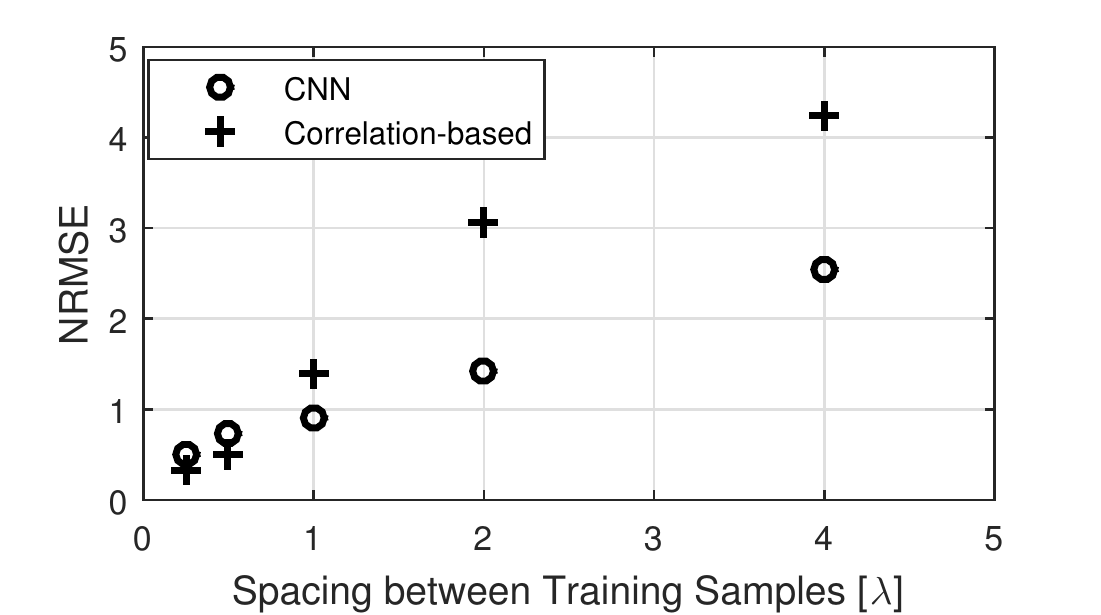}
    \caption{ NRMSE obtained by different positioning approaches for different spacings between samples of the uniform training grid.}
    \label{fig:NRMSEvsSpacing}
\end{figure}

To finalize, we analyze the impact of spatial sampling during training. For benchmarking, we contrast the CNN performance with the performance of the correlation-based classifier (\ref{eq:ML}).  We use  the CNN hyper-parameters that attained the lowest MSE in Fig. \ref{fig:MSEpos}, namely, the model with $L=4$ and $K=20$.\footnote{Ideally, the CNN hyper-parameters should be tuned according to the current training set. However, we keep the same hyper-parameterization throughout this analysis, for simplicity.} Fig. \ref{fig:NRMSEvsSpacing} contrasts the NRMSEs obtained from a CNN and the correlation-based classifier (\ref{eq:ML}), against spatial sampling in the training set. Overall, both approaches are able to attain fractional-wavelength accuracies at smaller training densities.  Noticeably, the CNN tend to behave better than (\ref{eq:ML}) for less dense training sampling. Given that (\ref{eq:ML}) does not  have interpolation abilities, this result is closely connected with the inherent interpolation abilities of the CNNs. The fact that the CNNs achieve similar, or even superior performance compared to standard non-parametric approaches while having attractive implementation complexity further corroborates their use in fingerprint-based localization systems.

%Mention that the size is dependent of the learning problem at hand. We choose a area of 25 by 25 meters and provided positioning results accordingly, for a proof-of-concept. Positioning in areas with different area is possible as long as the size of the network scales accordingly to the learning problem.

%See what happens when space is sampled density and density, and see how one has overfitting, but the test error increases.

% Mention that my results are much better than the ones in...

%Try to discretize the space into regions, and do classification. Then plot the classification error as a function of the number of regions.

\section{Takeaways and Further Work}

We have investigated a novel approach for massive MIMO fingerprint-based positioning by means of CNNs and measured channel snapshots. CNNs have a feedforward structure that is able to compactly summarize relevant positioning information in large channel data sets. The positioning capabilities of CNNs tend to generalize well, e.g. in highly-clustered propagation scenarios with or without LOS, thanks to their inherent feature learning abilities. Proper design allows fractional-wavelength positioning to be obtained under real-time requirements, and with low signal bandwidths.

The current investigation showcased some of the potentials of CNNs for positioning using channels with a complex structure. However, the design of CNNs in this contexts  should be a matter of further investigation, in order to be able to deal with  real-world impairments during the fingerprinting process. In this vein, some questions raised during this study are, for example, \textit{i)} how to achieve a robust CNN design that is able to deal with  impairments such as measurement and labeling noise, or channel variations that are not represented in the training set, or \textit{ii)} how to design complex-valued CNNs that perform well and are robust during optimization.

%\normalsize
%\small{ \bibliographystyle{IEEEtran}
%\bibliography{BIB}}


\end{document}